\documentclass{article}
\usepackage{spconf,amsmath,graphicx}

\usepackage{amsmath,bm,bbm,mathtools}
\usepackage{amsthm}
\usepackage[autostyle]{csquotes}
\usepackage{subcaption,multirow}
\usepackage{amssymb,pifont}
\usepackage{hyperref}       
\usepackage{url}            
\usepackage[scr=boondox]{mathalfa}
\usepackage{booktabs}       
\usepackage{amsfonts}       
\usepackage{nicefrac}       
\usepackage{microtype}      
\usepackage[colorinlistoftodos]{todonotes}
\usepackage{soul}
\usepackage{xcolor,colortbl}
\usepackage{booktabs}
\usepackage{algorithm}
\usepackage{algorithmic}

\definecolor{anti-flashwhite}{rgb}{0.95, 0.95, 0.96}
\definecolor{antiquewhite}{rgb}{0.98, 0.92, 0.84}

\newcolumntype{g}{>{\columncolor{gray}}c}
\newcolumntype{w}{>{\columncolor{white}}c}
\newcolumntype{d}{>{\columncolor{anti-flashwhite}}c}
\newcolumntype{a}{>{\columncolor{antiquewhite}}c}

\def\eqref#1{equation~\ref{#1}}

\def\1{\bm{1}}

\DeclareMathAlphabet{\mathsfit}{\encodingdefault}{\sfdefault}{m}{sl}
\SetMathAlphabet{\mathsfit}{bold}{\encodingdefault}{\sfdefault}{bx}{n}

\title{SEMI-SUPERVISED LEARNING FOR TEXT CLASSIFICATION \\ BY LAYER PARTITIONING}
\name{Alexander Hanbo Li, Abhinav Sethy}
\address{Alexa AI, Amazon}
\begin{document}
%
\maketitle
\begin{abstract}
Most recent neural semi-supervised learning algorithms rely on adding small perturbation to either the input vectors or their representations. These methods have been successful on computer vision tasks as the images form a continuous manifold, but are not appropriate for discrete input such as sentence. To adapt these methods to text input, we propose to decompose a neural network $M$ into two components $F$ and $U$ so that $M = U\circ F$. \footnote{``$\circ$" stands for composition, hence $U\circ F(x) = U(F(x))$.} The layers in $F$ are then frozen and only the layers in $U$ will be updated during most time of the training. In this way, $F$ serves as a feature extractor that maps the input to high-level representation and adds systematical noise using dropout. We can then train $U$ using any state-of-the-art SSL algorithms such as $\Pi$-model, temporal ensembling, mean teacher, etc. Furthermore, this gradually unfreezing schedule also prevents a pretrained model from catastrophic forgetting. The experimental results demonstrate that our approach provides improvements when compared to state of the art methods especially on short texts.
\end{abstract}
\section{Introduction}\label{sec:introduction}
Semi-supervised learning (SSL) \cite{chapelle2009semi} has been proved powerful for leveraging unlabeled data when we lack the resources to create large scale labeled dataset. In contrast with supervised learning methods that could only use labeled examples, SSL effectively uses the unlabeled samples to learn the underlying distribution of the data. Most SSL algorithms rely on an extra consistency or smoothness regularization which enforces the model to make consistent predictions on an input and its slightly perturbed version \cite{bachman2014learning,laine2016temporal,sajjadi2016regularization,tarvainen2017mean,miyato2018virtual,clark2018semi}. The perturbation, however, is always made by adding artificial noise (e.g. Gaussian noise) to the input $x$. In image classification tasks, the inputs are images that can be represented by dense vectors in a continuous space. However, in text classification tasks, each input text is a sequence of tokens and each token is represented by an one-hot vector which forms a sparse high-dimensional space. Even when we use word embeddings, the underlying input sentences themselves are still discrete. In this case, adding continuous noises to the discrete inputs become inappropriate. To tackle this problem, \cite{miyato2016adversarial} proposed to perturb each word by adding adversarial noise to the word embedding. However, it is hard to understand how independently adding noise to each token changes the sentence, as the perturbed embedding does not map back to any word. Ideally, after applying a perturbation function to a sentence, the noisy output should still represent a proper sentence and carries similar meaning.


Large-scale pretraining and transfer learning has achieved huge success on many NLP tasks \cite{mikolov2013distributed,dai2015semi,mou2016transferable,mccann2017learned,peters2017semi,peters2018deep,howard2018universal,devlin2018bert,radford2018improving}. The models are pretrained on some large dataset like Wikipedia's pages and then adapted to new domains by fine-tuning. For example, ULMFiT \cite{howard2018universal} divides the training procedure for text classification into three steps: (1) general domain language model (LM) pretraining, (2) target task LM fine-tuning and (3) target task classifier fine-tuning. Following these steps, they obtained state-of-the-art results on various text classification tasks. However, the fine-tuning step (2) seems redundant especially when the new-domain texts are short, and hence we ask ourselves whether the same level of performance can be achieved without LM fine-tuning.

To tackle the previous question, we propose one framework that allows adapting successful techniques in computer vision to NLP tasks. We avoid using artificial noises by decomposing a neural network into two  components $F$ and $U$ so that $M = U\circ F$, and freeze $F$ as both feature extractor and perturbation function. Because of the general-domain pretraining, layers in $F$ carry domain-agnostic knowledge while layers in $U$ are more task-specific. By training $U$ on the outputs of $F$, we can use any state-of-the-art SSL algorithm that depends on continuous input variations. The similar technique is also used in ULMFiT \cite{howard2018universal} to effectively prevent catastrophic forgetting. In the experiments, we combine the proposed layer partitioning (\textbf{LayerParti}) with $\Pi$-model or temporal ensembling (TE) \cite{laine2016temporal}, and test their performance to IMDB \cite{socher2013recursive} sentimental analysis and TREC-6 \cite{li2002learning} text classification. LayerParti achieves comparable result on IMDB and better result on TREC-6 compared with ULMFiT. Finally, we also apply our method to a spoken personal memory retrieval task and achieve better performance using only 6\% of the labels.

The paper is organized as follows. In Section \ref{sec:related}, we review some most relevant work. In Section \ref{sec:method}, we describe our method of layer partitioning for semi-supervised learning. In Section \ref{sec:experiments}, we evaluate the proposed method and discuss the results. In Section \ref{sec:conclusion}, we conclude by discussing our overall findings and contributions.

\section{Related Work}\label{sec:related}
In NLP, models pretrained on large-scale dataset \cite{peters2018deep,devlin2018bert,radford2018improving} learn useful general knowledge. One can hence transfer a pretrained model to new domains by fine-tuning it on domain-specific data. ULMFiT \cite{howard2018universal} standardizes the procedure into three steps: general domain language model (LM) pretraining, target task LM fine-tuning and target task classifier fine-tuning. It also gradually unfreezes the layers from top to bottom in order to prevent catastrophic forgetting. This idea is also used in computer vision, where one only fine-tunes the last layers \cite{donahue2014decaf,long2015fully} of a pretrained model. The common strategy is to use the lower layers to extract general features of the input.

For a comprehensive review of SSL methods, we refer readers to \cite{zhu2003semi,zhu2005semi} or \cite{oliver2018realistic}. We summarize four most relevant SSL methods below for continuous inputs. Our proposed method can be combined with any of them and applied to text classification. \\
\textbf{$\Pi$-Model}
The input and the prediction function in $\Pi$-Model are both stochastic, and hence it can produce different outputs for the same input $x$. $\Pi$-Model \cite{laine2016temporal,sajjadi2016regularization} adds a consistency loss which encourages the distance between a network's output for different passes of $x$ through the network to be small. However, the teacher network targets are very unstable and can change rapidly along the training. \\
\textbf{Temporal Ensembling and Mean Teacher}
\cite{laine2016temporal,tarvainen2017mean} proposed to obtain a more stable target by setting the teacher network as an exponential moving average of model parameters from previous training steps, or by directly calculating the moving average of previous targets. \\
\textbf{Virtual Adversarial Training}
Virtual Adversarial Training \cite{miyato2018virtual,miyato2016adversarial} approximates an adversarial noise to add to the input so that prediction function will be mostly affected. Note that in NLP, the noise is added to word embedding.

\section{Semi-Supervised Learning by Layer Partitioning}
\label{sec:method}

\begin{figure*}[htb]
    \small
    \centering
    \begin{subfigure}[b]{0.38\linewidth}
        \centering
        \includegraphics[width=0.95\linewidth]{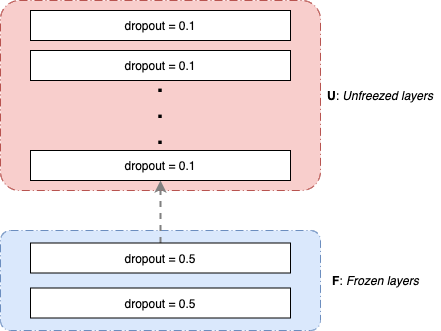}
        \caption{\label{fig:decomposition}Decomposition of a deep neural network.}
    \end{subfigure}
    ~
    \begin{subfigure}[b]{0.6\linewidth}
        \centering
        \includegraphics[width=1.0\linewidth]{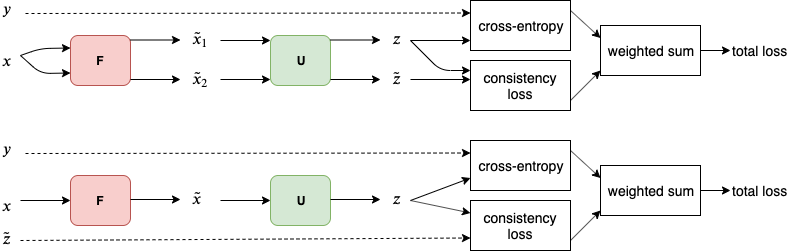}
        \caption{\label{fig:diag_model}LayerParti with $\Pi$-Model (top) and temporal ensembling (bottom). We use two colors red and green to distinguish between \textit{frozen} and \textit{unfrozen} parts.}
    \end{subfigure}
    \caption{The decomposition of a neural network and its application in SSL model training.}
\end{figure*}

\subsection{Partitioning Neural Network Layers}
Let $M$ be a neural network model with $n$ layers, we could then split $M$ into two \textbf{} parts $U$ and $F$, where $F$ contains the lower layers $\{1,\ldots,l\}$ and $U$ contains the higher layers $\{l+1,\ldots,n\}$. This is demonstrated in Figure \ref{fig:decomposition}. In a language model, the lower layers tend to learn more general knowledge \cite{devlin2018bert,howard2018universal} and are domain-agnostic. Therefore, we propose to \textbf{freeze} the layers in $F$ and use them as a feature extractor. We only update the task-specific layers in $U$. The similar strategy has been used in computer vision \cite{donahue2014decaf,long2015fully}. In fact,  assuming we have samples $x_1,\ldots,x_N$, freezing $F$ is equivalent to training the model $U$ with transformed inputs $F(x_1),\ldots,F(x_N)$. 

Depending on the splitting level $l$, the feature mapping $F$ may stand for the word embeddings or more abstract features containing complicated context information. In practice, we notice that freezing only the lowest layers gives better results than freezing intermediate or high layers. This is aligned with the intuition that higher layers learn task-specific features that cannot be directly transferred to new domains. In our experiments, we freeze the word embedding layer and the first layer of the transformer encoder.


\subsection{Perturbation of Textual Input}
$F$ can then be used to add systematical noise to the input by using dropout \cite{srivastava2014dropout}, as shown in Figure \ref{fig:decomposition}. Each time $x$ passes through $F$, the output will contain different perturbation. In another word, instead of adding artificial noise to $x$ to get $\tilde{x} \gets x+\epsilon$, we now have $\tilde{x} \gets F(x)$. Because the layers in $F$ are pretrained on a general domain, $F(x)$ is more likely to contain the same text information than $x + \epsilon$. For example, in sentimental analysis, changing one word \textit{happy} to \textit{sad} can totally change the sentiment of a sentence. This situation will rarely happen because of the language model property of $F$.

\subsection{Consistency Constraints}
Now being able to properly perturb the discrete input, we can use any state-of-the-art semi-supervised learning method to train $U$. We choose two methods in this paper: $\Pi$-Model and temporal ensembling (TE) \cite{laine2016temporal}.

The diagrams of the two models are shown in Figure \ref{fig:diag_model}. In $\Pi$-Model, each text $x$ is passed through the frozen layers $F$ twice to get two perturbed outputs $\tilde{x}_1$ and $\tilde{x}_2$, which are then fed into $U$ to get two predictions $z$ and $\tilde{z}$ of the class probabilities (i.e. $z_i \in \mathbb{R}^k$ where $k$ is the number of classes). The final loss is then a weighted sum of the cross-entropy loss (CE) and the consistency loss. Following \cite{laine2016temporal}, we use mean squared error (MSE) as the consistency loss, and hence the total loss $L(x,y) = CE(z, y) + w(t) \cdot MSE(z, \tilde{z})$, where $w(t)$ is the weight of the consistency loss and is a function of the iteration $t$.

The temporal ensembling model is similar to $\Pi$-Model, except that the ``teacher" targets $\tilde{z}$ is an ensemble of previous predictions. More formally, we update $Z_i \gets \alpha Z_i + (1-\alpha)z_i$ and then set the target as $\tilde{z}_i = Z_i / (1-\alpha^{T_i})$ where $T_i$ is the number of times that $x_i$ has been used. The factor $1-\alpha^{T_i}$ is to correct the zero initialization bias \cite{kingma2014adam,laine2016temporal}.

\subsection{Gradually Unfreezing}
Approaching the end of the training, we will gradually unfreeze the layers in $F$. The motivation is that $U$ has been well trained on $\{F(x)\}$ and becomes saturated, we can then unfreeze $F$ to let it also learn some specific features of the task domain. We describe the algorithm in detail in Algorithm \ref{alg:LayerParti}.

\begin{algorithm}[htb]
\small
\caption{The training procedure of LayerParti.}
\label{alg:LayerParti}
\begin{algorithmic}
    \STATE {\textbf{Input:} $\mathcal{D}_{L}$: labeled data, $\mathcal{D}_{UL}$: unlabeled data, $w(t)$: weight.}
    \STATE{\textbf{Initialize:} Frozen layers $F$, unfrozen layers $U$. In this paper, $F$ contains the word embedding and the first encoder layer. Set $z_i = 0$, $Z_i = 0$, $T_i = 0$ for all $i$.}
    \WHILE{$t \le$ max iterations}
        \STATE{$B \gets$ the new batch}
        \STATE{$\tilde{x}_i \gets F(x_i)$, $z_i \gets U(\tilde{x}_i)$ for all $i \in B$}
		\STATE{$L_{CE} \gets \frac{1}{|\mathcal{D}_L \cap B|}\sum_{i \in \mathcal{D}_L \cap B}\text{CE}(z_i, y_i)$}
		\IF{using $\Pi$-Model}
		\STATE{$\tilde{x}^{'}_i \gets F(x_i)$, $\tilde{z}_i \gets U(\tilde{x}^{'}_i)$}
		\ELSIF{using Temporal Ensembling}
		\STATE{$\tilde{z}_i \gets Z_i / (1 - \alpha^{T_i})$}
		\STATE{$T_i \gets T_i + 1$}
		\STATE{$Z_i \gets \alpha Z_i + (1-\alpha) z_i$}
		\ENDIF
		\STATE{$L_{consist} \gets \frac{1}{|\mathcal{D}_{UL} \cap B|}\sum_{i \in \mathcal{D}_{UL} \cap B} MSE(z_i, \tilde{z}_i)$}
		\STATE{$L \gets L_{CE} + w(t)L_{consist}$}
		\STATE{Back-propagate the gradients $\nabla L$ and update the layers in $U$.}
		\IF{$t \ge$ 80\% max iterations}
		\STATE{If $F$ is not empty, unfreeze the top layer in $F$ and put it to $U$.}
		\ENDIF
	\ENDWHILE
\end{algorithmic}
\end{algorithm}

\section{Experiments}\label{sec:experiments}
We test the proposed method \textit{LayerParti} with $\Pi$-Model or temporal ensembling (TE), but note that our method can be combined with other SSL algorithms like mean teacher \cite{tarvainen2017mean}, SWA \cite{izmailov2018averaging,athiwaratkun2018there}, FGE \cite{garipov2018loss}, etc. 

\subsection{Public Datasets}
IMDB \cite{maas2011learning} dataset consists of movie reviews from the Internet Movie Database (IMDb) labeled as positive or negative. The TREC-6 dataset \cite{li2002learning} consists of open-domain, fact-based questions divided into six semantic categories. We summarize the statistics of the datasets in Table \ref{tab:dataset}. For both datasets, we truncate the texts to not exceed 256 tokens.

\begin{table}[htb]
    \small
    \centering
    \begin{tabular}{cccccc}
        \toprule
               & Train  & Test   & Unlabeled & Avg & Max  \\
        \midrule
         IMDB  & 25,000 & 25,000 & 50,000    & 239 & 2,506 \\
         TREC-6&  5,452 &    500 & -         & 10  & 45 \\ 
         \bottomrule
    \end{tabular}
    \caption{Statistics of the datasets. ``Train" and ``Test" mean the number of labeled data in training and test partitions respectively. ``Avg" and ``Max" refer to the average and maximum text length.}
    \label{tab:dataset}
\end{table}

\subsection{Experimental Settings}
We use a Transformer encoder from \cite{wolf2019transformers} with 16 layers, 410 dimensional embedding, 2100 dimensional hidden layer and 10 heads for each multi-head attention layer.  The encoder was pretrained with a linear language model heading on the WikiText-103 \cite{merity2016pointer} data. We also use the BERT-tokenizer from \cite{wolf2019transformers} where a [CLS] token is appended to each sentence. Then we simply add a linear classification layer on top of the embedding of the [CLS] token to predict the class.

\begin{figure}[htb]
    \small
    \centering
    \includegraphics[width=0.99\linewidth]{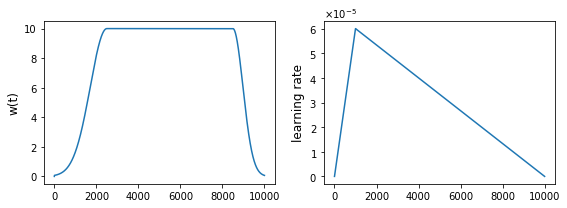}
    \caption{The schedule of $w(t)$ with max value 10. We use the same scheduling function as in \cite{laine2016temporal}.}
    \label{fig:schedule}
\end{figure}

For supervised learning baselines, we train the model for 3 epochs with batch size 16 and we also clip the gradient norm to be less than 0.4. We choose Adam \cite{kingma2014adam} optimizer with default parameter settings. The learning rate is triangular as shown in Figure \ref{fig:schedule} with warm-up proportion as 10\%. The weight $w(t)$ has the same scheduling function as in \cite{laine2016temporal} with 25\% of the iterations for warm-up and 15\% for ramp-down. With unlabeled data, we instead train each model for 8 epochs with 4 gradient accumulation steps. We sample the data so that 25\% of each batch are labeled. The dropout rate in $F$ is 0.5 for $\Pi$-Model and 0.3 for temporal ensembling, and the dropout rate in $U$ is always 0.1.

\subsection{Quantitative Results}

From Table \ref{tab:real_data_results}, LayerParti combined with $\Pi$-Model or TE provide the same level of results compared with SOTA \cite{nlpprogress}. On IMDB (long texts), our results are comparable to supervised ULMFiT results in \cite{howard2018universal} where they fine-tuned both the language model and the classifier on supervised data. But we do notice that their semi-supervised results (fine-tuning LM on all data and training classifier on labeled ones) are better than ours. However, on TREC-6, with all the labels available, our method acts as a regularization method and achieves 97.2\% test accuracy that is even better than the best semi-supervised ULMFiT result of 96.4\%. Also with only 60 labels, we can produce better accuracy than the semi-supervised ULMFiT results \cite{howard2018universal} with 100 labels. 

The previous analysis suggests that on short texts, our method is very competitive and can give better accuracy. But on longer texts like IMDB reviews, fine-tuning the LM on all the data (not just on supervised ones) is indeed beneficial.

\begin{table}[htb]
    \small
    \centering
    \begin{tabular}{cc}
        \toprule
                           &  Accuracy \\
        \midrule
        IMDB(all)*                          &  90.5\% \\
        IMDB(500)                        & 81.5\% \\
        IMDB(100)                         & 64.7\% \\
        \midrule
        IMDB(all+unsup) + LayerParti ($\Pi$)      &  91.4\% \\
        IMDB(500 + unsup) + LayerParti ($\Pi$)    &  83.2\% \\
        IMDB(100 + unsup) + LayerParti ($\Pi$)    &  69.3\% \\
        \midrule
        IMDB(all+unsup) + LayerParti (TE)      & \textbf{91.8\%} \\
        IMDB(500 + unsup) + LayerParti (TE)    &  \textbf{85.1\%} \\
        IMDB(100 + unsup) + LayerParti (TE)    &  \textbf{75.9\%} \\
        \midrule
        \midrule
        TREC-6(all)             &  95.8\% \\
        TREC-6(400)        &  81.8\% \\
        TREC-6(60)      & 44.6\%\\
        \midrule
        TREC-6(all) +  LayerParti ($\Pi$) & \textbf{97.2\%}  \\
        TREC-6(400) + LayerParti ($\Pi$) & 85.8\% \\
        TREC-6(60) + LayerParti ($\Pi$) & 65.4\% \\
        \midrule
        TREC-6(all) +  LayerParti (TE) & 96.4\%  \\
        TREC-6(400) + LayerParti (TE) & \textbf{86.0\%} \\
        TREC-6(60) + LayerParti (TE) & \textbf{67.8\%} \\
        \bottomrule
    \end{tabular}
    \caption{Accuracy results on IMDB and TREC-6 text classifications. *The IMDB baseline using all the data is worse than the 94.1\% accuracy of VAT \cite{miyato2018virtual} but they are using 400 max length for text truncation while ours is 256.}
    \label{tab:real_data_results}
\end{table}

\subsection{Personal Memory Retrieval}
\begin{figure}[htb]
    \centering
    \small
    \includegraphics[width=0.8\linewidth]{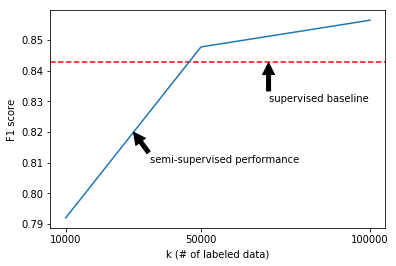}
    \caption{Comparison of F1 scores. The \textit{supervised baseline} is trained on all the examples. To get \textit{semi-supervised} performance, we trained the models on $k$ labeled data (half positive and half negative) and masked out the label information of the rest.}
    \label{fig:pds}
\end{figure}

\noindent Recent progress of machine learning has largely enriched the functionality of smart personal assistants, including allowing users to store and retrieve long-term personal memory \cite{fakoor2018direct}. We apply our method to such a task where the memory is created by voice input and converted to text using a speech recognition system. The queries are also spoken and transcribed automatically.

On this task, we change the model to have two encoder layers, 300 dimensional embedding, 512 dimensional hidden layer and 5 heads. The full training dataset contains 871K labeled examples among which 52K are positive and 819K are negative. For supervised training, we use a sampler to guarantee each batch contains balanced positive and negative examples. The average text length is 7 and we truncate all the input at length 10. The test data contains 93K examples with 3K positives and 90K negatives. The results are in Figure \ref{fig:pds}. We observe that our method achieves better F1 using only 50,000 labeled data which is less than 6\% of the labels used by the supervised baseline.

\section{Conclusion}\label{sec:conclusion}
We proposed a semi-supervised learning framework (LayerParti) for discrete text input. When combined with $\Pi$-Model or temporal ensembling -- both proposed for image inputs -- our method achieves competitive results on three text classification tasks. Especially when dealing with short texts like TREC-6 or personal memory retrieval, our framework provides better performance without LM fine-tuning.

\clearpage
\vfill\pagebreak
\bibliographystyle{IEEEbib}
\bibliography{refs}

\begin{thebibliography}{10}

\bibitem{chapelle2009semi}
Olivier Chapelle, Bernhard Scholkopf, and Alexander Zien,
\newblock ``Semi-supervised learning (chapelle, o. et al., eds.; 2006)[book
  reviews],''
\newblock {\em IEEE Transactions on Neural Networks}, vol. 20, no. 3, pp.
  542--542, 2009.

\bibitem{bachman2014learning}
Philip Bachman, Ouais Alsharif, and Doina Precup,
\newblock ``Learning with pseudo-ensembles,''
\newblock in {\em Advances in Neural Information Processing Systems}, 2014, pp.
  3365--3373.

\bibitem{laine2016temporal}
Samuli Laine and Timo Aila,
\newblock ``Temporal ensembling for semi-supervised learning,''
\newblock {\em arXiv preprint arXiv:1610.02242}, 2016.

\bibitem{sajjadi2016regularization}
Mehdi Sajjadi, Mehran Javanmardi, and Tolga Tasdizen,
\newblock ``Regularization with stochastic transformations and perturbations
  for deep semi-supervised learning,''
\newblock in {\em Advances in Neural Information Processing Systems}, 2016, pp.
  1163--1171.

\bibitem{tarvainen2017mean}
Antti Tarvainen and Harri Valpola,
\newblock ``Mean teachers are better role models: Weight-averaged consistency
  targets improve semi-supervised deep learning results,''
\newblock in {\em Advances in neural information processing systems}, 2017, pp.
  1195--1204.

\bibitem{miyato2018virtual}
Takeru Miyato, Shin-ichi Maeda, Masanori Koyama, and Shin Ishii,
\newblock ``Virtual adversarial training: a regularization method for
  supervised and semi-supervised learning,''
\newblock {\em IEEE transactions on pattern analysis and machine intelligence},
  vol. 41, no. 8, pp. 1979--1993, 2018.

\bibitem{clark2018semi}
Kevin Clark, Minh-Thang Luong, Christopher~D Manning, and Quoc~V Le,
\newblock ``Semi-supervised sequence modeling with cross-view training,''
\newblock {\em arXiv preprint arXiv:1809.08370}, 2018.

\bibitem{miyato2016adversarial}
Takeru Miyato, Andrew~M Dai, and Ian Goodfellow,
\newblock ``Adversarial training methods for semi-supervised text
  classification,''
\newblock {\em arXiv preprint arXiv:1605.07725}, 2016.

\bibitem{mikolov2013distributed}
Tomas Mikolov, Ilya Sutskever, Kai Chen, Greg~S Corrado, and Jeff Dean,
\newblock ``Distributed representations of words and phrases and their
  compositionality,''
\newblock in {\em Advances in neural information processing systems}, 2013, pp.
  3111--3119.

\bibitem{dai2015semi}
Andrew~M Dai and Quoc~V Le,
\newblock ``Semi-supervised sequence learning,''
\newblock in {\em Advances in neural information processing systems}, 2015, pp.
  3079--3087.

\bibitem{mou2016transferable}
Lili Mou, Zhao Meng, Rui Yan, Ge~Li, Yan Xu, Lu~Zhang, and Zhi Jin,
\newblock ``How transferable are neural networks in nlp applications?,''
\newblock {\em arXiv preprint arXiv:1603.06111}, 2016.

\bibitem{mccann2017learned}
Bryan McCann, James Bradbury, Caiming Xiong, and Richard Socher,
\newblock ``Learned in translation: Contextualized word vectors,''
\newblock in {\em Advances in Neural Information Processing Systems}, 2017, pp.
  6294--6305.

\bibitem{peters2017semi}
Matthew~E Peters, Waleed Ammar, Chandra Bhagavatula, and Russell Power,
\newblock ``Semi-supervised sequence tagging with bidirectional language
  models,''
\newblock {\em arXiv preprint arXiv:1705.00108}, 2017.

\bibitem{peters2018deep}
Matthew~E Peters, Mark Neumann, Mohit Iyyer, Matt Gardner, Christopher Clark,
  Kenton Lee, and Luke Zettlemoyer,
\newblock ``Deep contextualized word representations,''
\newblock {\em arXiv preprint arXiv:1802.05365}, 2018.

\bibitem{howard2018universal}
Jeremy Howard and Sebastian Ruder,
\newblock ``Universal language model fine-tuning for text classification,''
\newblock {\em arXiv preprint arXiv:1801.06146}, 2018.

\bibitem{devlin2018bert}
Jacob Devlin, Ming-Wei Chang, Kenton Lee, and Kristina Toutanova,
\newblock ``Bert: Pre-training of deep bidirectional transformers for language
  understanding,''
\newblock {\em arXiv preprint arXiv:1810.04805}, 2018.

\bibitem{radford2018improving}
Alec Radford, Karthik Narasimhan, Tim Salimans, and Ilya Sutskever,
\newblock ``Improving language understanding by generative pre-training,''
\newblock {\em URL https://s3-us-west-2. amazonaws.
  com/openai-assets/researchcovers/languageunsupervised/language understanding
  paper. pdf}, 2018.

\bibitem{socher2013recursive}
Richard Socher, Alex Perelygin, Jean Wu, Jason Chuang, Christopher~D Manning,
  Andrew Ng, and Christopher Potts,
\newblock ``Recursive deep models for semantic compositionality over a
  sentiment treebank,''
\newblock in {\em Proceedings of the 2013 conference on empirical methods in
  natural language processing}, 2013, pp. 1631--1642.

\bibitem{li2002learning}
Xin Li and Dan Roth,
\newblock ``Learning question classifiers,''
\newblock in {\em Proceedings of the 19th international conference on
  Computational linguistics-Volume 1}. Association for Computational
  Linguistics, 2002, pp. 1--7.

\bibitem{donahue2014decaf}
Jeff Donahue, Yangqing Jia, Oriol Vinyals, Judy Hoffman, Ning Zhang, Eric
  Tzeng, and Trevor Darrell,
\newblock ``Decaf: A deep convolutional activation feature for generic visual
  recognition,''
\newblock in {\em International conference on machine learning}, 2014, pp.
  647--655.

\bibitem{long2015fully}
Jonathan Long, Evan Shelhamer, and Trevor Darrell,
\newblock ``Fully convolutional networks for semantic segmentation,''
\newblock in {\em Proceedings of the IEEE conference on computer vision and
  pattern recognition}, 2015, pp. 3431--3440.

\bibitem{zhu2003semi}
Xiaojin Zhu, Zoubin Ghahramani, and John~D Lafferty,
\newblock ``Semi-supervised learning using gaussian fields and harmonic
  functions,''
\newblock in {\em Proceedings of the 20th International conference on Machine
  learning (ICML-03)}, 2003, pp. 912--919.

\bibitem{zhu2005semi}
Xiaojin~Jerry Zhu,
\newblock ``Semi-supervised learning literature survey,''
\newblock Tech. {R}ep., University of Wisconsin-Madison Department of Computer
  Sciences, 2005.

\bibitem{oliver2018realistic}
Avital Oliver, Augustus Odena, Colin~A Raffel, Ekin~Dogus Cubuk, and Ian
  Goodfellow,
\newblock ``Realistic evaluation of deep semi-supervised learning algorithms,''
\newblock in {\em Advances in Neural Information Processing Systems}, 2018, pp.
  3235--3246.

\bibitem{srivastava2014dropout}
Nitish Srivastava, Geoffrey Hinton, Alex Krizhevsky, Ilya Sutskever, and Ruslan
  Salakhutdinov,
\newblock ``Dropout: a simple way to prevent neural networks from
  overfitting,''
\newblock {\em The journal of machine learning research}, vol. 15, no. 1, pp.
  1929--1958, 2014.

\bibitem{kingma2014adam}
Diederik~P Kingma and Jimmy Ba,
\newblock ``Adam: A method for stochastic optimization,''
\newblock {\em arXiv preprint arXiv:1412.6980}, 2014.

\bibitem{izmailov2018averaging}
Pavel Izmailov, Dmitrii Podoprikhin, Timur Garipov, Dmitry Vetrov, and
  Andrew~Gordon Wilson,
\newblock ``Averaging weights leads to wider optima and better
  generalization,''
\newblock {\em arXiv preprint arXiv:1803.05407}, 2018.

\bibitem{athiwaratkun2018there}
Ben Athiwaratkun, Marc Finzi, Pavel Izmailov, and Andrew~Gordon Wilson,
\newblock ``There are many consistent explanations of unlabeled data: Why you
  should average,''
\newblock {\em arXiv preprint arXiv:1806.05594}, 2018.

\bibitem{garipov2018loss}
Timur Garipov, Pavel Izmailov, Dmitrii Podoprikhin, Dmitry~P Vetrov, and
  Andrew~G Wilson,
\newblock ``Loss surfaces, mode connectivity, and fast ensembling of dnns,''
\newblock in {\em Advances in Neural Information Processing Systems}, 2018, pp.
  8789--8798.

\bibitem{maas2011learning}
Andrew~L Maas, Raymond~E Daly, Peter~T Pham, Dan Huang, Andrew~Y Ng, and
  Christopher Potts,
\newblock ``Learning word vectors for sentiment analysis,''
\newblock in {\em Proceedings of the 49th annual meeting of the association for
  computational linguistics: Human language technologies-volume 1}. Association
  for Computational Linguistics, 2011, pp. 142--150.

\bibitem{wolf2019transformers}
Thomas Wolf, Lysandre Debut, Victor Sanh, Julien Chaumond, Clement Delangue,
  Anthony Moi, Pierric Cistac, Tim Rault, Rémi Louf, Morgan Funtowicz, and
  Jamie Brew,
\newblock ``Transformers: State-of-the-art natural language processing,'' 2019.

\bibitem{merity2016pointer}
Stephen Merity, Caiming Xiong, James Bradbury, and Richard Socher,
\newblock ``Pointer sentinel mixture models,''
\newblock {\em arXiv preprint arXiv:1609.07843}, 2016.

\bibitem{nlpprogress}
``{NLP}-progress,'' \url{https://nlpprogress.com/},
\newblock Repository to track the progress in NLP, including the datasets and
  the current state-of-the-art for the most common NLP tasks.

\bibitem{fakoor2018direct}
Rasool Fakoor, Amanjit Kainth, Siamak Shakeri, Christopher Winestock,
  Abdel-rahman Mohamed, and Ruhi Sarikaya,
\newblock ``Direct optimization of f-measure for retrieval-based personal
  question answering,''
\newblock in {\em 2018 IEEE Spoken Language Technology Workshop (SLT)}. IEEE,
  2018, pp. 815--822.

\end{thebibliography}

\end{document}